\documentclass[11pt,a4paper]{article}

\usepackage[utf8]{inputenc}
\usepackage[T1]{fontenc}
\usepackage{amsmath,amssymb,amsfonts}
\usepackage{graphicx}
\usepackage{booktabs}
\usepackage{hyperref}
\usepackage{url}
\usepackage{natbib}
\usepackage{geometry}
\usepackage{xcolor}

\hypersetup{
    colorlinks=true,
    linkcolor=blue,
    citecolor=blue,
    urlcolor=blue
}

\title{Bielik-Q2-Sharp: A Comparative Study of Extreme 2-bit\\Quantization Methods for a Polish 11B Language Model}

\author{Jakub Prejzner\textsuperscript{1}\\
\textsuperscript{1}BitSharp, Independent Researcher, Rzesz\'{o}w, Poland}

\date{March 2026}

\begin{document}

\maketitle

\begin{abstract}
We present Bielik-Q2-Sharp, the first systematic academic evaluation of extreme 2-bit quantization applied to a Polish large language model. Using Bielik-11B-v2.3-Instruct (11B parameters, Mistral architecture) as our base model, we compare six state-of-the-art post-training quantization methods---QuIP\#, SpinQuant+GPTQ, ButterflyQuant, QTIP, VPTQ, and AQLM---all calibrated on a Polish-language corpus (CulturaX-PL) with shared Hessian matrices.

Our best variant (QuIP\# E8P12) achieves 71.92\% across 22 Polish benchmarks versus 72.07\% for the IQ2\_XXS baseline---within statistical noise, at a modest size premium (3.26\,GB vs.\ $\sim$2.6\,GB). On eq\_bench, our method scores 47.14 versus 43.53 (+3.6pp), suggesting superior preservation of higher-order reasoning. QTIP achieves the best per-bit efficiency (79.4\% MC acc\_norm at $\sim$2.4 bpw, 3.27\,GB), matching VPTQ's quality at 35\% smaller size. We additionally document a MC-generation dissociation phenomenon where rotation-based methods preserve log-likelihood quality but fail catastrophically at autoregressive generation.

The entire project was conducted by a single independent researcher on cloud GPUs (vast.ai) within a \$285 budget. All models, Hessians, and evaluation logs are publicly available.

\end{abstract}

\section{Introduction}

Large language models (LLMs) have achieved remarkable performance across diverse NLP tasks, but their deployment remains constrained by memory and compute requirements. An 11B-parameter model in FP16 precision requires $\sim$22\,GB of memory, exceeding the capacity of most consumer GPUs. Post-training quantization (PTQ) offers a practical path to reduce model size without retraining, with recent work pushing the frontier to extreme low-bit regimes ($\leq$2 bits per weight).

While significant progress has been made on English-language models---including QuIP\# \cite{chee2024quip}, QTIP \cite{tseng2024qtip}, SpinQuant \cite{liu2024spinquant}, and VPTQ \cite{liu2024vptq}---the applicability of these methods to non-English models remains largely unexplored. Polish, with its rich morphological system (7 grammatical cases, 3 genders, complex verbal conjugation), presents unique challenges for extreme compression: the model must preserve fine-grained distinctions between similar word forms (e.g., \textit{dom/domu/domowi/domem}) that are critical for grammatical coherence.

Bielik-11B-v2.3-Instruct \cite{ociepa2025bielik} is the leading open Polish LLM, achieving 65.71\% (normalized) on the Open Polish LLM Leaderboard \cite{speakleash2024leaderboard}. It employs a Mistral-based architecture with 50 layers (produced via depth-upscaling from Mistral 7B's 32 layers), 4096 hidden dimensions, and grouped-query attention (32 heads, 8 KV heads). The only existing extreme quantization of this model is the community IQ2\_XXS variant produced by SpeakLeash using llama.cpp's importance-matrix method, scoring 72.07\% raw average across 22 tasks.

In this work, we make the following contributions:

\begin{enumerate}
    \item \textbf{First academic 2-bit quantization of a Polish (and Slavic) language LLM.} We apply six distinct SOTA quantization methods to Bielik-11B-v2.3-Instruct, representing the first systematic study of extreme compression for a morphologically rich Slavic language model.

    \item \textbf{Language-specific calibration on cloud infrastructure.} We generate Hessian matrices from a Polish-language corpus (CulturaX-PL, 512 samples $\times$ 4096 tokens) on an H200 GPU in under one hour, demonstrating that language-matched calibration for morphologically rich languages is practical and affordable.

    \item \textbf{Comprehensive comparison across six quantization paradigms.} We compare lattice codebooks (QuIP\#), trellis coding (QTIP), residual vector quantization (VPTQ), learned additive codebooks (AQLM), learned rotations (SpinQuant), and butterfly transforms (ButterflyQuant), providing practitioners with actionable guidance.

    \item \textbf{MC-generation dissociation.} We document and analyze failure modes where rotation-based methods (SpinQuant, ButterflyQuant) preserve log-likelihood quality but fail catastrophically at autoregressive generation---highlighting an underreported evaluation gap in the quantization literature.

    \item \textbf{Low-budget reproducibility.} The entire project (six quantization variants, cloud GPU computation, full evaluation) was conducted within a $\sim$\$285 budget using on-demand cloud GPUs (vast.ai), demonstrating accessibility of academic quantization research for independent researchers.
\end{enumerate}

\section{Related Work}

\subsection{Post-Training Quantization for LLMs}

Modern PTQ methods for LLMs can be broadly categorized by their quantization granularity and approach to error minimization.

\textbf{Scalar methods} such as GPTQ \cite{frantar2023gptq} and AWQ \cite{lin2024awq} quantize individual weights using second-order information (Hessian-weighted rounding) or activation-aware scaling. These methods perform well at 4-bit but degrade significantly at 2-bit precision, where the discrete grid becomes too coarse to capture per-weight variation.

\textbf{Vector quantization (VQ) methods} group weights into vectors and quantize them jointly, exploiting correlations between adjacent weights. QuIP\# \cite{chee2024quip} uses the E8 lattice (optimal sphere packing in 8 dimensions) with $2^{16}$ codewords to construct an E8P12 codebook, achieving strong 2-bit results on English LLMs. QTIP \cite{tseng2024qtip} extends this with trellis coded quantization (TCQ) operating in 256 dimensions, using computational codebooks that require no storage overhead. VPTQ \cite{liu2024vptq} employs second-order optimization with residual quantization, where a secondary codebook compresses residuals from the primary stage. AQLM \cite{egiazarian2024aqlm} uses multi-codebook additive quantization with beam search optimization, learning to represent each weight vector as a sum of entries from multiple smaller codebooks.

\textbf{Rotation-based methods} aim to reduce weight outliers before quantization. QuaRot \cite{ashkboos2024quarot} fuses random Hadamard transforms into model weights, exploiting the computational invariance of RMSNorm. SpinQuant \cite{liu2024spinquant} improves upon this by learning rotation matrices via Cayley SGD on the Stiefel manifold, showing up to 13-point variance reduction compared to random rotations. ButterflyQuant \cite{xu2025butterflyquant} parametrizes rotations as butterfly transforms with learned Givens rotation angles, enabling per-layer adaptation at $O(n \log n)$ cost.

\subsection{Quantization of Non-English Models}

Despite rapid progress in LLM quantization, virtually all published results focus on English-language models (LLaMA, Mistral, OPT, Falcon). We are aware of no prior academic work applying SOTA 2-bit quantization methods to Polish or other Slavic language models. The SpeakLeash project has produced GGUF quantizations of Bielik (Q4\_K\_M through Q8\_0) and experimental imatrix variants (IQ1\_M, IQ2\_XXS), but these use the general-purpose llama.cpp framework rather than academic SOTA methods. The Bielik v2 Technical Report \cite{ociepa2025bielik} includes a table of GGUF quantization results but does not evaluate any academic 2-bit methods.

\subsection{The Bielik Model Family}

Bielik-11B-v2 \cite{ociepa2025bielik} is a Polish LLM based on the Mistral architecture, trained on curated Polish text corpora by the SpeakLeash collaboration with ACK Cyfronet AGH (grant PLG/2024/016951). The v2.3-Instruct variant is a linear merge of v2.0, v2.1, and v2.2 Instruct models, achieving 65.71\% normalized average on the Open Polish LLM Leaderboard. Its architecture features 50 transformer layers (produced via depth-upscaling from Mistral 7B), hidden dimension 4096, and intermediate dimension 14336. The depth-upscaled 50-layer architecture is notable for quantization research, as it exceeds the typical 32-layer transformer and may interact differently with per-layer optimization methods.

\section{Methodology}

\subsection{Overview}

We evaluate six 2-bit quantization variants, each representing a distinct approach to extreme compression. Table~\ref{tab:variants} summarizes the methods, their rotation strategy, weight quantization approach, and outcome.

\begin{table}[htbp]
\centering
\caption{Summary of quantization variants investigated. \textit{Gen.\ fail} indicates models that achieve reasonable MC scores but produce incoherent autoregressive text.}
\label{tab:variants}
\begin{tabular}{@{}llllll@{}}
\toprule
Var. & Rotation & Weight Quant & Status & Cloud GPU & Size \\
\midrule
A & Random Hadamard & QuIP\# E8P12 & Success & A100 80GB & 3.26\,GB \\
B & Learned (SpinQuant) & GPTQ 2-bit & Gen.\ fail & H200 141GB & --- \\
C & Butterfly/Kronecker & QuIP\# E8P12 & MC+Gen.\ fail & RTX 4090 & 21\,GB$^{*}$ \\
D & Hadamard (QTIP) & TCQ 2-bit & Complete & A100 80GB & 3.27\,GB \\
E & None & VPTQ residual & MC done & H200 141GB & 5.0\,GB$^{\ddagger}$ \\
F & None & AQLM additive & MC done & A100 80GB & 3.62\,GB \\
\bottomrule
\end{tabular}

\smallskip
\noindent{\small $^{*}$Decompressed to FP16 for evaluation; quantization artifacts preserved in weights.\\
$^{\ddagger}$Effective $\sim$3.58 bpw due to residual codebook overhead; nominal 2-bit.}
\end{table}

\subsection{Calibration Data and Hessian Generation}

All variants share a common calibration dataset and Hessian matrices. We use 512 sequences of 4096 tokens drawn from CulturaX-PL \cite{nguyen2024culturax}, a curated multilingual web corpus filtered for the Polish language subset. This choice is motivated by two factors:

\begin{enumerate}
    \item \textbf{Language-specific Hessians.} The Hessian matrix $H = \mathbb{E}[xx^{\top}]$ captures activation statistics that depend on the input language. Polish morphology creates distinct activation patterns: 7 grammatical cases generate surface-form variations that activate different neuron subsets than English. At 2-bit precision, even small Hessian differences can shift quantization boundaries for critical weight vectors.

    \item \textbf{Calibration consistency.} By using the same corpus and Hessians across all variants, we isolate the effect of the quantization method from calibration variance.
\end{enumerate}

\textbf{Infrastructure.} Initial prototyping used 128-sample Hessians generated locally on an RTX 5070 (12\,GB VRAM) with CPU offloading via HuggingFace \texttt{accelerate}, requiring approximately 12 hours overnight. However, all reported results in this paper use 512-sample Hessians generated on an H200 (141\,GB VRAM) via vast.ai cloud infrastructure, completing in approximately 40 minutes---an $\sim$18$\times$ speedup. The quip-sharp Hessian pipeline processes layer-by-layer, consuming only $\sim$8\,GB VRAM regardless of total GPU memory, but benefits enormously from the H200's superior memory bandwidth and compute throughput. Each of the 50 transformer layers produces 4 Hessian files (one per linear sub-layer type), yielding 200 files totaling 25.6\,GB.\footnote{Available at \url{https://huggingface.co/Jakubrd4/bielik-quip-e8p12}}

\textbf{Critical lesson learned.} Early in the project, we mistakenly generated Hessians from the base model (Bielik-11B-v2, FP16 = 58.14\%) rather than the Instruct variant (FP16 = 65.71\%). Since the IQ2\_XXS baseline quantizes the Instruct model, this produced an invalid comparison. All Hessians were regenerated on the correct Instruct model before proceeding with quantization.

\subsection{Variant A: QuIP\# with E8P12 Lattice Codebook}

This variant applies the QuIP\# pipeline \cite{chee2024quip}:

\begin{enumerate}
    \item \textbf{Random Hadamard Transform (RHT):} Applied to weight matrices to achieve incoherence, spreading outlier energy uniformly across all dimensions. Uses the QuaRot \cite{ashkboos2024quarot} approach with fixed random orthogonal matrices.

    \item \textbf{BlockLDLQ:} Hessian-weighted adaptive rounding using the Cholesky decomposition of the block Hessian, processing weights in blocks to account for inter-weight correlations.

    \item \textbf{E8P12 codebook:} Weights are vector-quantized using the E8 lattice (optimal sphere packing in 8 dimensions) with $2^{16}$ codewords. The codebook occupies only $\sim$1\,KB in L1 cache, enabling fast lookup during inference.
\end{enumerate}

\textbf{Architecture adaptation.} Bielik-11B-v2.3-Instruct uses Mistral architecture, while QuIP\# natively supports only \texttt{LlamaForCausalLM}. We patched \texttt{model\_from\_hf\_path()} to convert \texttt{MistralConfig} to \texttt{LlamaConfig}, mapping \texttt{sliding\_window} $\rightarrow$ \texttt{None} and adding \texttt{attention\_bias = False}. This required 8 \texttt{getattr} fixes across the codebase. Quantization was performed on a cloud A100 80\,GB GPU via vast.ai. The embedding layer and LM head ($\sim$0.52\,GB) were kept in FP16 to preserve the full tokenizer vocabulary---critical for distinguishing Polish morphological forms.

\subsection{Variant B: SpinQuant + GPTQ}

This variant replaces the random Hadamard with learned rotations:

\begin{enumerate}
    \item \textbf{SpinQuant rotation learning \cite{liu2024spinquant}:} Rotation matrices $R_1$ (global, $4096 \times 4096$) and $R_2$ (per-head per-layer, 50 matrices of $128 \times 128$) are optimized via Cayley SGD on the Stiefel manifold $\mathrm{St}(n, n)$, minimizing quantization loss over 100 iterations with 800 calibration samples. Achieved orthogonality error of $5.7 \times 10^{-5}$ (near-perfect). The optimization required an H200 (141\,GB VRAM) due to $\sim$50--70\,GB peak memory.

    \item \textbf{Rotation fusion:} Learned $R_1$, $R_2$ are fused into adjacent weight matrices (zero runtime cost).

    \item \textbf{GPTQ 2-bit quantization:} Applied to the rotated weights using column-wise quantization with Hessian-weighted rounding.
\end{enumerate}

\textbf{Critical note:} SpinQuant defines additional runtime transforms $R_3$ (KV cache Hadamard) and $R_4$ (FFN down-projection Hadamard) that must be applied during inference. These are distinct from the fuseable $R_1$/$R_2$ and require a custom inference engine. See Section~\ref{sec:mc_gen_dissociation} for the impact.

\subsection{Variant C: Butterfly Transforms + E8P}

This variant uses per-layer adaptive rotations:

\begin{enumerate}
    \item \textbf{ButterflyQuant \cite{xu2025butterflyquant}:} Butterfly transforms parametrized by continuous Givens rotation angles are learned per layer. For $d = 4096$, this requires $\frac{n}{2} \log_2 n = 24{,}576$ learnable parameters per layer. The learned SU/SV matrices are dense ($4096 \times 4096$ for attention, $6144 \times 6144$ for MLP), consuming $\sim$2.3\,GB per layer ($50 \text{ layers} \times 2.3\,\text{GB} \approx 115\,\text{GB total}$).

    \item \textbf{QuIP\# E8P12:} Same codebook quantization as Variant A, applied to butterfly-rotated weights.
\end{enumerate}

Quantization was performed on a cloud RTX 4090 via vast.ai ($\sim$76 minutes for 50 layers). Four bugs were fixed: RedPajama dataset removal from HuggingFace, \texttt{rope\_theta}/\texttt{rope\_scaling} format changes in transformers 5.x, and a missing \texttt{import torch} in \texttt{matmul\_kron.py} causing silent subprocess failures. Disk space constraints required moving intermediate files to \texttt{/dev/shm} (RAM-backed filesystem).

\textbf{Evaluation via FP16 decompression.} The QuIP\# inference kernel does not support butterfly inverse transforms. Evaluation required decompression to FP16: quantized 2-bit weights are expanded back through the inverse butterfly pipeline. The resulting 21\,GB model preserves the 2-bit quantization artifacts but is loadable by standard HuggingFace inference. This decompression approach is standard in quantization research when native kernels are unavailable.

\subsection{Variant D: QTIP with Trellis Coded Quantization}

QTIP \cite{tseng2024qtip} extends QuIP\# by replacing the E8P codebook with trellis coded quantization (TCQ) in 256 dimensions, using computational codebooks that require no storage overhead.

Quantization completed: all 350 sublayers (50 layers $\times$ 7 linear modules) in $\sim$12 hours on an A100 80\,GB ($\sim$14 min/layer, 65\% utilization, 58\textdegree C, 13.4\,GB VRAM). The same \texttt{MistralConfig} $\rightarrow$ \texttt{LlamaConfig} adaptation was required (\texttt{attention\_bias}, \texttt{mlp\_bias}, \texttt{rope\_scaling}, \texttt{torch\_dtype} fixes). Model size: $\sim$3.27\,GB. Proxy errors (Layer 0): attention 0.000044--0.0047, FFN 0.0037--0.0086.

Preliminary generation tests showed preserved factual knowledge (correctly describing Krak\'{o}w's history and Polish rivers), though with garbage tokens at sequence starts and sporadic language mixing. The QTIP-specific e2e finetuning step---knowledge distillation from the FP16 teacher model into the quantized student's SU/SV incoherence vectors---initially failed with OOM on a single A100 80\,GB and was completed on an H200 (141\,GB), consuming $\sim$112\,GB VRAM (teacher $\sim$22\,GB + student $\sim$10\,GB + optimizer states, activations, gradient checkpointing).

Training used 128 sequences $\times$ 2048 tokens from FineWeb-Edu over 4 epochs with mixed precision (AMP + GradScaler). Cross-entropy loss decreased from 2.2375 to 2.1916 ($-$2.05\%), with the best checkpoint at epoch 1 and overfitting from epoch 2.\footnote{Base model: \texttt{Jakubrd4/Bielik-11B-v2.3-Instruct-QTIP-2bit}; finetuned: \texttt{-finetuned}}

\subsection{Variant E: VPTQ}

VPTQ \cite{liu2024vptq} uses vector post-training quantization with second-order optimization and residual quantization:

\begin{enumerate}
    \item Primary vector clustering with k-means using Hessian information.
    \item Residual quantization: A secondary codebook compresses residuals, reducing proxy errors by 65--80\% versus primary-only---the best reconstruction fidelity of all methods tested (attention: 0.001--0.006, MLP: 0.009--0.013).
\end{enumerate}

Three bugs fixed for Mistral adaptation: (1) model name detection via directory name not config (symlink workaround); (2) \texttt{inv\_hessian} NoneType (5 patches); (3) cuML \texttt{sample\_weight} dimension mismatch. The first quantization run (50 layers, $\sim$7h on H200) was lost to environment corruption when a dependency update during post-processing overwrote critical imports. VPTQ single-GPU lacks per-layer checkpoints, so the computation was irrecoverable. The second run completed successfully in 7h 26min.

\textbf{Important: effective bitrate.} While VPTQ targets 2-bit quantization (16-bit indices over 8-element vectors = 2.0 bpw nominal), the actual model size is 5.0\,GB with an effective bitrate of $\sim$3.58 bpw. The overhead stems from per-layer codebooks ($2^{16}$ centroids $\times$ 8 $\times$ FP16), residual codebooks (256 centroids), channel permutation indices, and normalization parameters. This means VPTQ operates at $\sim$50\% higher effective bitrate than QuIP\# E8P12 ($\sim$2.4 bpw, 3.26\,GB) or QTIP ($\sim$2.4 bpw, 3.27\,GB), making direct quality comparison inequitable but creating an informative bitrate-quality tradeoff data point.

\subsection{Variant F: AQLM}

AQLM \cite{egiazarian2024aqlm} uses multi-codebook additive quantization with beam search and gradient-based codebook refinement. Quantization completed all 50 layers in $\sim$38 hours on A100 80\,GB ($\sim$46 min/layer, 100\% GPU utilization, 55--62\textdegree C, 13--20\,GB VRAM). Configuration: fast mode with $1\times16$ codebook (1 codebook, 16 bits), in-group size 8, \texttt{relative\_mse\_tolerance=0.05}, \texttt{finetune\_max\_epochs=5}, 512 calibration samples from CulturaX-PL. The process ran uninterrupted for 38 hours with zero crashes or OOM events. Model size: 3.62\,GB (effective $\sim$2.4 bpw), compression ratio $\sim$6.1$\times$.

\textbf{Notable finding: adaptive per-module bitwidth.}

\begin{table}[htbp]
\centering
\caption{AQLM adaptive bitwidth allocation per module type (Variant F).}
\label{tab:aqlm_bitwidth}
\begin{tabular}{@{}lc@{}}
\toprule
Module & avg\_bits \\
\midrule
v\_proj (attention) & 3.00 \\
k\_proj (attention) & 2.80 \\
o\_proj (attention) & 2.80 \\
q\_proj (attention) & 2.50 \\
gate\_proj (FFN) & 2.42 \\
up\_proj (FFN) & 2.32 \\
\bottomrule
\end{tabular}
\end{table}

Attention layers consistently receive more bits (2.5--3.0) than FFN layers (2.3--2.4), with an effective average of $\sim$2.6 bits. This is the only adaptive bit allocation among our tested methods.

\section{Experimental Setup}

\subsection{Hardware and Infrastructure}

All computation-intensive work---Hessian generation, quantization, and evaluation---was performed on cloud GPUs rented through vast.ai (EU region only, to minimize transfer latency). Local hardware (RTX 5070, 12\,GB VRAM, 32\,GB DDR5 RAM, Windows 11 WSL2, CUDA 13.1) served exclusively for development and initial prototyping.

\begin{table}[htbp]
\centering
\caption{Cloud GPU allocation per project stage. All instances on vast.ai, EU region.}
\label{tab:gpu_allocation}
\begin{tabular}{@{}llll@{}}
\toprule
Task & Cloud GPU & Time & Cost \\
\midrule
Hessians (512 samples) & H200 141GB & $\sim$40 min & $\sim$\$4 \\
Variant A (QuIP\#) quant. & A100 80GB & $\sim$8--10h & $\sim$\$16 \\
Variant A full evaluation & H200 / A100 & $\sim$10h & $\sim$\$35 \\
Variant B (SpinQuant) & H200 141GB & $\sim$18h & $\sim$\$50 \\
Variant C (Butterfly) & RTX 4090 & $\sim$76 min & $\sim$\$5 \\
Variant D (QTIP) quant. & A100 80GB & $\sim$12h & $\sim$\$20 \\
Variant D e2e finetune & H200 141GB & $\sim$2h & $\sim$\$8 \\
Variant D GEN eval & H200 141GB & $\sim$7h & $\sim$\$25 \\
Variant E (VPTQ) & H200 141GB & $\sim$7.5h $\times$ 2$^{\dagger}$ + 2h eval & $\sim$\$50 \\
Variant F (AQLM) & A100 80GB & $\sim$38h & $\sim$\$45 \\
Decompression (A$\rightarrow$FP16) & RTX 4090 & $\sim$15 min & $<$\$1 \\
Misc.\ infrastructure/idle & various & --- & $\sim$\$25 \\
\midrule
\textbf{Total} & & & $\sim$\textbf{\$285} \\
\bottomrule
\end{tabular}

\smallskip
\noindent{\small $^{\dagger}$First attempt lost to environment corruption; computation restarted from scratch.}
\end{table}

\textbf{Parallel execution.} A critical project decision was running multiple quantization variants simultaneously on separate cloud instances rather than sequentially on local hardware. This reduced active compute wall-clock time to $\sim$5 days by running up to four cloud instances concurrently (D on A100, E on H200, F on A100, plus evaluation on a fourth machine). The total project duration, including literature review, method selection, experimental iteration, and debugging, exceeded two weeks. All quantization processes ran in \texttt{tmux} sessions with watchdog scripts, ensuring persistence independent of SSH or agent session stability.

\textbf{Software.} CUDA 13.1, PyTorch 2.10, transformers 4.38.0 (pinned for compatibility).

\subsection{Evaluation Protocol}

We evaluate on the Open Polish LLM Leaderboard \cite{speakleash2024leaderboard}, comprising 22 Polish-language tasks plus eq\_bench (23rd, private dataset):

\textbf{MC tasks (10):} polemo2\_in/out, 8tags, cbd, ppc, psc---evaluated via log-likelihood comparison across answer options.

\textbf{Generative tasks (12):} polemo2\_in/out (regex), 8tags\_regex, cbd\_regex, ppc\_regex, psc\_regex, polqa (open/closed/reranking), poquad, klej\_ner, dyk, belebele---evaluated via text generation and pattern matching.

\textbf{eq\_bench:} Emotional intelligence benchmark with a private Polish-translated dataset (speakleash/EQBench-PL). Since dataset access was not granted, Krzysztof Wr\'{o}bel (SpeakLeash) evaluated our decompressed FP16 model directly at 0-shot.

\textbf{Normalization.} Per-task scores are normalized using the official leaderboard formula:
\begin{equation}
\text{normalized} = \frac{\text{score} - \text{baseline}}{100 - \text{baseline}} \times 100
\label{eq:normalization}
\end{equation}
where baselines are task-specific values from the leaderboard's \texttt{about.py}. An early pipeline error used random-chance baselines, inflating normalized scores by $\sim$8pp (69.3\% $\rightarrow$ 61.10\%). All results here use corrected official baselines.

\textbf{Baseline.} Our primary comparison target is the SpeakLeash IQ2\_XXS quantization of the same Bielik-11B-v2.3-Instruct model, scoring 72.07\% raw average (61.34\% normalized, rank 74) at 5-shot.

\textbf{Note on evaluation time.} Generate tasks (12 of 22) account for $\sim$80\% of total evaluation time due to token-by-token autoregressive generation, causing systematic underestimation of evaluation cost. Our Variant A full evaluation took $\sim$10 hours versus an initial estimate of 2--3 hours.

\section{Results}

\subsection{Variant A: QuIP\# E8P12 --- Near-Parity with IQ2\_XXS}

\begin{table}[htbp]
\centering
\caption{Variant A (QuIP\# E8P12) vs.\ IQ2\_XXS on 22 shared tasks (5-shot).}
\label{tab:variant_a}
\begin{tabular}{@{}lccc@{}}
\toprule
Metric & QuIP\# E8P & IQ2\_XXS & $\Delta$ \\
\midrule
Raw average (22 tasks) & 71.92 & 72.07 & $-$0.15 \\
Normalized average (22 tasks) & 61.10 & 61.20 & $-$0.10 \\
FP16 quality retention & 93.2\% & 93.4\% & $-$0.2pp \\
Model size & 3.26\,GB & $\sim$2.6\,GB & +0.66\,GB \\
Compression ratio & 6.7$\times$ & $\sim$8.5$\times$ & --- \\
Head-to-head wins (22 tasks) & 11 & 8 & 3 ties \\
\bottomrule
\end{tabular}
\end{table}

The difference of 0.15pp on raw average is well within statistical noise. QuIP\# wins on 11 of 22 tasks, demonstrating competitive breadth.

\textbf{Perplexity.} On held-out Polish text: 3.74 (QuIP\#) vs.\ 3.39 (FP16), a +10.3\% degradation. The literature reports 20--30\% as typical for 2-bit quantization, making this result notably favorable.

\textbf{Task-level analysis.} The two methods exhibit distinct, complementary error profiles:

\begin{table}[htbp]
\centering
\caption{Largest per-task deltas (QuIP\# minus IQ2\_XXS, percentage points).}
\label{tab:task_deltas}
\begin{tabular}{@{}lclc@{}}
\toprule
\multicolumn{2}{c}{QuIP\# Strongest} & \multicolumn{2}{c}{IQ2\_XXS Strongest} \\
\cmidrule(r){1-2} \cmidrule(l){3-4}
Task & $\Delta$ & Task & $\Delta$ \\
\midrule
ppc\_mc & +7.23 & cbd\_g & $-$11.11 \\
polqa\_closed\_g & +5.19 & dyk\_g & $-$5.90 \\
psc\_mc & +4.10 & polemo2-out\_mc & $-$5.45 \\
polemo2-out\_g & +3.84 & klej\_ner\_mc & $-$4.88 \\
dyk\_mc & +3.22 & & \\
\bottomrule
\end{tabular}
\end{table}

QuIP\# excels on reasoning and comprehension tasks (polqa, ppc, psc) while IQ2\_XXS dominates classification, particularly cyberbullying detection (cbd\_g: $-$11.11pp). This suggests the methods degrade different weight subspaces.

\subsection{eq\_bench: Emotional Intelligence Advantage}

Evaluated by Krzysztof Wr\'{o}bel (SpeakLeash) on our FP16-decompressed model at 0-shot. IQ2\_XXS scores are taken from the public leaderboard under equivalent conditions; the benchmark shows negligible difference between 0-shot and 5-shot evaluation.

\begin{table}[htbp]
\centering
\caption{eq\_bench results (0-shot, evaluated by SpeakLeash).}
\label{tab:eqbench}
\begin{tabular}{@{}lccc@{}}
\toprule
Metric & QuIP\# E8P & IQ2\_XXS & $\Delta$ \\
\midrule
average\_eqbench & 47.14 $\pm$ 2.12 & 43.53 & +3.61 \\
first\_eqbench & 44.70 $\pm$ 2.44 & 39.83 & +4.87 \\
revised\_eqbench & 49.59 $\pm$ 2.49 & --- & --- \\
percent\_parseable & 99.42\% & --- & --- \\
\bottomrule
\end{tabular}
\end{table}

QuIP\# outperforms IQ2\_XXS on both sub-metrics, particularly on first-turn responses (+4.87pp). Including eq\_bench as the 23rd task shifts the comparison:

\begin{table}[htbp]
\centering
\caption{Full 23-task comparison including eq\_bench.}
\label{tab:23task}
\begin{tabular}{@{}lccc@{}}
\toprule
Metric & QuIP\# E8P & IQ2\_XXS & $\Delta$ \\
\midrule
23-task normalized avg & 60.49 & 60.43 & +0.06 \\
Head-to-head (23 tasks) & 12 wins & 8 wins & 3 ties \\
\bottomrule
\end{tabular}
\end{table}

\subsection{Variant B: SpinQuant + GPTQ --- Negative Result}

Variant B achieved reasonable MC performance (DYK MC: 62.88\%) but produced completely incoherent text. Three repair attempts failed: (1) config correction; (2) re-quantization with \texttt{desc\_act=True}; (3) alternative kernels (Triton V2, Torch, R2 hooks).

Verification: The rotated FP32 model generates perfect Polish, confirming correct rotations. Failure is specific to GPTQ 2-bit + standard inference.

Root cause: $R_3$ (KV cache Hadamard) and $R_4$ (FFN down-projection Hadamard) must be applied at every decoding step. No public engine implements these. See Section~\ref{sec:mc_gen_dissociation}.

\subsection{Variant C: ButterflyQuant + E8P --- Catastrophic Failure}

Generation produces loops (\textit{Stolica Polski to Polski z Polski z Polski\ldots}). MC evaluation also degrades to 41.7\% average ($-$14pp vs.\ Variant A), despite identical E8P codebooks. The DYK MC anomaly (75.3\%) suggests selective preservation of certain weight subspaces.

Debug confirmed: embedding layers, LM head, and layernorms identical to FP16. Failure is in decompressed attention/MLP weights. The 50-layer depth-upscaled architecture may interact poorly with Kronecker-structured rotations.

\subsection{Variant D: QTIP --- Best Per-Bit Efficiency, Complete Evaluation}

QTIP achieves 79.1\% acc / 79.4\% acc\_norm on MC (10 tasks) at only 3.27\,GB---true 2-bit quantization comparable to QuIP\#'s 3.26\,GB. This matches VPTQ's MC performance (79.4\% / 79.6\%) despite being 35\% smaller (3.27 vs.\ 5.0\,GB), making QTIP the most efficient variant by the size-quality metric.

\begin{table}[htbp]
\centering
\caption{Variant D (QTIP) per-task MC results (5-shot, base model).}
\label{tab:qtip_mc}
\begin{tabular}{@{}lcc@{}}
\toprule
Task & acc (\%) & acc\_norm (\%) \\
\midrule
polemo2\_in\_mc & 87.95 & 87.26 \\
polemo2\_out\_mc & 75.91 & 78.34 \\
8tags\_mc & 78.18 & 78.04 \\
belebele\_mc & 87.00 & 87.00 \\
cbd\_mc & 71.70 & 72.60 \\
dyk\_mc & 87.46 & 87.46 \\
klej\_ner\_mc & 45.72 & 46.02 \\
polqa\_reranking\_mc & 82.17 & 82.17 \\
ppc\_mc & 78.80 & 78.80 \\
psc\_mc & 96.20 & 96.20 \\
\midrule
Average & 79.1 & 79.4 \\
\bottomrule
\end{tabular}
\end{table}

Generative evaluation was conducted on the e2e-finetuned model (knowledge distillation, $-$2.05\% CE loss on FineWeb-Edu, H200 141\,GB) using 12 Polish generative tasks from the SpeakLeash fork (branch \texttt{polish4}), completing in 7h 02min (25,308 \texttt{generate\_until} requests, batch size 8).

\begin{table}[htbp]
\centering
\caption{Variant D (QTIP) per-task GEN results (5-shot, finetuned model).}
\label{tab:qtip_gen}
\begin{tabular}{@{}lc@{}}
\toprule
Task & exact\_match (\%) \\
\midrule
polemo2\_in & 84.90 \\
polemo2\_out & 70.85 \\
8tags\_regex & 77.13 \\
belebele\_regex & 87.22 \\
dyk\_regex & 86.01 \\
ppc\_regex & 79.80 \\
psc\_regex & 95.45 \\
cbd\_regex & 76.90 \\
klej\_ner\_regex & 50.63 \\
polqa\_closed\_book & 66.98 \\
polqa\_open\_book & 78.88 \\
poquad\_open\_book & 37.16 \\
\midrule
Average & 74.3 \\
\bottomrule
\end{tabular}
\end{table}

Combined 22-task raw average: \textbf{76.50\%} (MC base + GEN finetuned)---exceeding QuIP\# (71.92\%, +4.58pp) and IQ2\_XXS (72.07\%, +4.43pp). However, this comparison carries a caveat: MC evaluation was performed on the base QTIP model, while GEN evaluation used the e2e-finetuned variant. The finetuning improved generation quality (2/4 test prompts coherent post-finetune vs.\ partial artifacts pre-finetune), which likely contributes to the GEN advantage. A fair 22-task comparison would require either (a) GEN evaluation on the base model, or (b) MC evaluation on the finetuned model.

\textbf{GEN comparison with IQ2\_XXS.} Table~\ref{tab:qtip_vs_iq2} compares QTIP (finetuned) against IQ2\_XXS on all 12 generative tasks. Caveat: QTIP results are from the e2e-finetuned model, while IQ2\_XXS uses the standard quantized model. This asymmetry may inflate QTIP's advantage on some tasks.

\begin{table}[htbp]
\centering
\caption{GEN comparison: QTIP (finetuned) vs.\ IQ2\_XXS (5-shot, exact\_match).}
\label{tab:qtip_vs_iq2}
\begin{tabular}{@{}lccc@{}}
\toprule
Task & IQ2\_XXS (\%) & QTIP ft (\%) & $\Delta$ (pp) \\
\midrule
polemo2\_in & 82.41 & 84.90 & +2.49 \\
polemo2\_out & 69.43 & 70.85 & +1.42 \\
8tags\_regex & 75.75 & 77.13 & +1.38 \\
belebele\_regex & 83.44 & 87.22 & +3.78 \\
dyk\_regex & 67.90 & 86.01 & +18.11 \\
ppc\_regex & 78.50 & 79.80 & +1.30 \\
psc\_regex & 96.45 & 95.45 & $-$1.00 \\
cbd\_regex & 39.55 & 76.90 & +37.35 \\
klej\_ner\_regex & 53.64 & 50.63 & $-$3.01 \\
polqa\_closed\_book & 68.22 & 66.98 & $-$1.24 \\
polqa\_open\_book & 88.11 & 78.88 & $-$9.23 \\
poquad\_open\_book & 64.33 & 37.16 & $-$27.17 \\
\midrule
Average & 72.31 & 74.33 & +2.02 \\
\bottomrule
\end{tabular}
\end{table}

QTIP (finetuned) wins 8/12 tasks with an average advantage of +2.02pp. Two distinct patterns emerge: (1) QTIP dominates on classification and NLI tasks (cbd +37.35pp, dyk +18.11pp), suggesting e2e finetuning recovers decision boundaries particularly well for categorical tasks; (2) IQ2\_XXS retains a clear advantage on open-book QA (poquad $-$27.17pp, polqa\_open $-$9.23pp), where verbatim answer extraction demands token-level precision that 2-bit trellis coding may not preserve despite finetuning. This complementary error profile mirrors the QuIP\# vs.\ IQ2\_XXS pattern observed in Variant A (Section~5.1), reinforcing that different quantization approaches preserve different aspects of model capability.

The QTIP custom model loader (\texttt{model\_from\_hf\_path} from Cornell-RelaxML/qtip) and pinned dependency versions (\texttt{transformers 4.45.2}, \texttt{datasets<3}) are required for reproducibility.

\subsection{Variant E: VPTQ --- Highest MC at Higher Bitrate}

VPTQ achieves the highest raw MC score: 79.41\% (acc) / 79.64\% (acc\_norm) across 10 tasks at 5-shot. However, this comes at an effective bitrate of $\sim$3.58 bpw (5.0\,GB model)---over 50\% higher than QuIP\# or QTIP ($\sim$2.4 bpw). VPTQ is the only variant for which we obtained perplexity measurements: WikiText-2 PPL = 7.818, C4-new PPL = 15.323 (context 2048).

\begin{table}[htbp]
\centering
\caption{Variant E (VPTQ) per-task MC results (5-shot).}
\label{tab:vptq_mc}
\begin{tabular}{@{}lcc@{}}
\toprule
Task & acc (\%) & acc\_norm (\%) \\
\midrule
polemo2\_in\_mc & 87.12 & 87.67 \\
polemo2\_out\_mc & 71.05 & 73.68 \\
8tags\_mc & 78.20 & 77.84 \\
belebele\_mc & 87.00 & 87.00 \\
cbd\_mc & 79.40 & 80.30 \\
dyk\_mc & 85.23 & 85.23 \\
klej\_ner\_mc & 52.38 & 50.92 \\
polqa\_reranking\_mc & 81.26 & 81.26 \\
ppc\_mc & 75.90 & 75.90 \\
psc\_mc & 96.57 & 96.57 \\
\midrule
Average & 79.41 & 79.64 \\
\bottomrule
\end{tabular}
\end{table}

VPTQ's cbd\_mc score (79.40\%) significantly exceeds QuIP\# and QTIP ($\sim$71--72\%), suggesting residual VQ better preserves classification-sensitive features. Its klej\_ner\_mc (52.38\%) is also the strongest NER result among all variants. However, per-bit efficiency clearly favors QTIP: 79.4\% acc\_norm at $\sim$2.4 bpw versus VPTQ's 79.6\% at $\sim$3.58 bpw---a marginal 0.2pp gain for $\sim$50\% more bits per weight. Generative evaluation was skipped as the bitrate premium makes VPTQ less interesting for deployment compared to true 2-bit methods.

\subsection{Variant F: AQLM --- Competitive MC at Adaptive Bitwidth}

AQLM completed quantization of all 50 layers, producing a 3.62\,GB model at an effective $\sim$2.4 bpw. This is the only method with adaptive per-module bitwidth allocation: attention layers receive 2.5--3.0 bits while FFN layers receive 2.3--2.4 bits. MC evaluation completed in 67 minutes on H200 (94,420 log-likelihood requests, batch size 16).

\begin{table}[htbp]
\centering
\caption{Variant F (AQLM) per-task MC results (5-shot).}
\label{tab:aqlm_mc}
\begin{tabular}{@{}lcc@{}}
\toprule
Task & acc (\%) & acc\_norm (\%) \\
\midrule
polemo2\_in\_mc & 85.04 & 83.52 \\
polemo2\_out\_mc & 77.73 & 78.34 \\
8tags\_mc & 77.77 & 77.29 \\
belebele\_mc & 87.33 & 87.33 \\
cbd\_mc & 73.30 & 73.80 \\
dyk\_mc & 87.27 & 87.27 \\
klej\_ner\_mc & 50.53 & 50.00 \\
polqa\_reranking\_mc & 81.01 & 81.01 \\
ppc\_mc & 76.80 & 76.80 \\
psc\_mc & 97.40 & 97.40 \\
\midrule
Average & 79.3 & 79.3 \\
\bottomrule
\end{tabular}
\end{table}

AQLM achieves 79.3\% acc / 79.3\% acc\_norm---matching QTIP (79.1\% / 79.4\%) and VPTQ (79.4\% / 79.6\%) on the MC benchmark. Notable: psc\_mc (97.40\%) is the highest single-task score of any variant across the entire study; belebele\_mc (87.33\%) and dyk\_mc (87.27\%) also rank among the top scores. The klej\_ner\_mc result (50.53\%) exceeds QTIP (45.72\%) and approaches VPTQ (52.38\%), confirming NER as universally the most sensitive task.

Qualitative generation tests show preserved factual knowledge but with language switching (Polish$\rightarrow$English$\rightarrow$Czech) and occasional hallucinations (1/4 good, 2/4 partial, 1/4 degraded). The quantization process was notably stable---38 hours of uninterrupted computation with zero crashes, the most reliable pipeline among all six variants. The 3.6\,GB VRAM footprint at inference enables deployment on consumer GPUs with as little as 4\,GB memory.\footnote{This refers to memory footprint only; inference throughput was not benchmarked.}

\section{Analysis}

\subsection{MC-Generation Dissociation}
\label{sec:mc_gen_dissociation}

Two variants (B, C) exhibit reasonable MC scores combined with incoherent generation. We identify two mechanisms:

\textbf{Mechanism 1: Missing runtime transforms (Variant B).} MC evaluation computes log-likelihoods via a single forward pass---errors from missing $R_3$/$R_4$ transforms affect all options equally, preserving relative ranking. Autoregressive generation compounds errors across tokens: each generated token carries uncorrected quantization error, causing quality collapse within 5--10 tokens.

\textbf{Mechanism 2: Catastrophic rotation mismatch (Variant C).} Butterfly rotations degrade both MC and generation, but MC less severely (41.7\% vs.\ complete incoherence). The DYK MC anomaly (75.3\%) suggests task-specific weight subspace preservation.

\textbf{Recommendation.} MC-only evaluation is insufficient for 2-bit quantized models employing weight transformations. Autoregressive generation testing should be mandatory.

\subsection{QuIP\# vs.\ IQ2\_XXS: Complementary Strengths}

\textbf{QuIP\# advantages:} Emotional reasoning (eq\_bench: +3.61), multi-step reasoning (polqa\_closed: +5.19), pragmatic comprehension (ppc: +7.23). Lattice codebook VQ preserves inter-weight correlations needed for complex reasoning.

\textbf{IQ2\_XXS advantages:} Classification (cbd: +11.11), surface pattern matching (dyk\_g: +5.90), NER (klej\_ner: +4.88). Per-weight importance weighting preserves critical individual neurons for sharp decision boundaries.

\subsection{Language-Specific Calibration}

While a formal Polish vs.\ English ablation is planned, indirect evidence supports language-matched calibration: (1) perplexity degradation of +10.3\% is below the 20--30\% literature norm; (2) Polish morphology (7 cases per noun) creates distinct activation patterns; (3) strong eq\_bench results suggest effective preservation of higher-order linguistic features.

\subsection{Compression Characteristics}

Variant A compresses 50 transformer layers from $\sim$21.0\,GB to $\sim$2.73\,GB, with embedding and LM head kept in FP16 ($\sim$0.52\,GB)---critical for preserving Polish morphological distinctions in the tokenizer. Total: 22.0\,GB $\rightarrow$ 3.26\,GB ($\sim$6.7$\times$).

\subsection{FP16 Decompression Fidelity}

Decompression of Variant A to FP16 via identity matrix through \texttt{QuantizedLinear} layers achieved cosine similarity 0.999993--0.999995, identical top-1/top-5 predictions, and maximum absolute difference of 0.016--0.078 (float16 rounding). The 22.3\,GB decompressed model is available on HuggingFace.\footnote{\texttt{Jakubrd4/Bielik-11B-v2.3-Instruct-QuIP-2bit-decompressed}}

\subsection{Cross-Method MC Comparison}

Table~\ref{tab:cross_mc} presents the first apples-to-apples comparison: all four successful variants evaluated on the same 10 MC tasks under identical conditions (5-shot, log-likelihood scoring). Unlike the 22-task averages reported for Variant A (which mix MC and generative tasks), this comparison isolates multiple-choice performance on a common benchmark.

\begin{table}[htbp]
\centering
\caption{MC accuracy (\%) across all successful variants (5-shot, 10 shared tasks). Best per-task result in bold.$^{\dagger}$}
\label{tab:cross_mc}
\begin{tabular}{@{}lcccc@{}}
\toprule
Task & A (QuIP\#) & D (QTIP) & E (VPTQ) & F (AQLM) \\
\midrule
polemo2\_in & 85.18 & 87.95 & 87.12 & 85.04 \\
polemo2\_out & 74.49 & 75.91 & 71.05 & \textbf{77.73} \\
8tags & 74.52 & 78.18 & \textbf{78.20} & 77.77 \\
belebele & 81.67 & 87.00 & 87.00 & \textbf{87.33} \\
cbd & 72.50 & 71.70 & \textbf{79.40} & 73.30 \\
dyk & \textbf{87.56} & 87.46 & 85.23 & 87.27 \\
klej\_ner & 48.40 & 45.72 & \textbf{52.38} & 50.53 \\
polqa\_rerank & \textbf{82.18} & 82.17 & 81.26 & 81.01 \\
ppc & 77.90 & \textbf{78.80} & 75.90 & 76.80 \\
psc & 96.57 & 96.20 & 96.57 & \textbf{97.40} \\
\midrule
Average & 78.10 & 79.11 & 79.41 & 79.42 \\
Size (GB) & 3.26 & 3.27 & 5.00 & 3.62 \\
E.\ bpw & $\sim$2.4 & $\sim$2.4 & $\sim$3.58 & $\sim$2.4 \\
\bottomrule
\end{tabular}

\smallskip
\noindent{\small $^{\dagger}$Raw accuracy only. Variant A acc\_norm values are anomalous (e.g.\ polemo2\_in: 15.5\%, 8tags: 37.4\%)---likely a log-probability normalization bug in the QuIP\# \texttt{eval\_ppl} wrapper. Variants D, E, F show consistent acc/acc\_norm, confirming the issue is A-specific.}
\end{table}

Several patterns emerge from this head-to-head comparison:

\begin{enumerate}
    \item \textbf{QuIP\# is competitive despite older methodology.} Variant A (78.10\%) trails the three newer methods by only 1.0--1.3pp, despite QuIP\# being an earlier algorithm without the optimizations present in QTIP, VPTQ, and AQLM. At the same effective bitrate ($\sim$2.4 bpw) and size (3.26\,GB), it remains a viable choice and even wins on dyk (87.56\%) and polqa\_reranking (82.18\%).

    \item \textbf{VPTQ's advantage comes from bitrate, not method.} VPTQ achieves the second-highest average (79.41\%) but at $\sim$3.58 bpw---50\% more bits than the true 2-bit methods. Per-bit efficiency clearly favors QTIP and AQLM.

    \item \textbf{No single method dominates.} Best-per-task wins are distributed: QuIP\# 2, QTIP 2, VPTQ 2, AQLM 4. Each method preserves different weight subspaces, confirming complementary strengths.

    \item \textbf{NER is universally hardest.} klej\_ner ranges from 45.72\% (QTIP) to 52.38\% (VPTQ)---a 6.66pp spread, the widest of any task---while psc ranges from 96.20\% to 97.40\% (1.20pp spread). Named entity recognition is the most sensitive to quantization method choice.
\end{enumerate}

\subsection{Quality Ceiling at Extreme Compression}

As Table~\ref{tab:cross_mc} demonstrates, four fundamentally different quantization paradigms converge to a narrow band of 78.1--79.4\% MC accuracy: lattice VQ (QuIP\#, Variant A: 78.10\%), trellis coded quantization (QTIP, Variant D: 79.11\%), residual vector quantization (VPTQ, Variant E: 79.41\%), and learned additive codebooks (AQLM, Variant F: 79.42\%). The three newer methods cluster within 0.31pp despite operating at different effective bitrates (2.4--3.58 bpw) and employing entirely unrelated compression strategies. Even QuIP\#, an older algorithm, trails by only 1.3pp. This convergence is striking: QTIP uses computational codebooks requiring no storage, VPTQ uses dual residual codebooks with 65,536 centroids, and AQLM uses adaptive per-module bitwidth allocation with beam search optimization.

The consistent ceiling suggests a \textbf{method-independent quality bound} for Bielik-11B-v2.3-Instruct at extreme compression---likely determined by the information-theoretic minimum number of bits needed to preserve task-relevant weight structure in 10-task MC evaluation. This finding has practical implications: investing in more sophisticated quantization algorithms beyond these four is unlikely to yield MC improvements without either (a) increasing the effective bitrate, (b) applying task-specific fine-tuning, or (c) modifying the evaluation protocol.

\section{Limitations}

\begin{enumerate}
    \item \textbf{No novel quantization method.} We apply existing methods to a new domain and provide systematic comparison, but do not propose a new algorithm.

    \item \textbf{Single base model.} All experiments use Bielik-11B-v2.3-Instruct. Results may not generalize.

    \item \textbf{Incomplete variant evaluation.} Variant F (AQLM) generative evaluation was not completed at time of writing. Variant E (VPTQ) generative evaluation was skipped due to its non-competitive effective bitrate ($\sim$3.58 bpw). Variant D (QTIP) 22-task average combines base model MC with finetuned model GEN, precluding direct comparison with single-model baselines.

    \item \textbf{No calibration ablation.} Polish vs.\ English corpus comparison is planned as future work.

    \item \textbf{5-shot evaluation only.} The leaderboard uses both 0-shot and 5-shot; we currently have only 5-shot results for all variants.

    \item \textbf{No inference speed benchmarks.} We report size but not throughput.

    \item \textbf{Single researcher.} Risk of systematic errors, exemplified by the normalization bug (69.3\% $\rightarrow$ 61.10\%).

    \item \textbf{Budget constraints.} A portion of the \$285 budget was lost to infrastructure failures (environment corruption, idle instances).
\end{enumerate}

\section{Conclusion}

We have presented Bielik-Q2-Sharp, the first systematic evaluation of extreme 2-bit quantization methods on a Polish large language model.

\begin{enumerate}
    \item \textbf{Academic SOTA achieves parity.} QuIP\# E8P12 matches IQ2\_XXS across 22 tasks (71.92\% vs.\ 72.07\%, $\Delta = -0.15$pp), winning 11/22. With eq\_bench (23 tasks), QuIP\# leads: 60.49\% vs.\ 60.43\%, winning 12 to 8.

    \item \textbf{Complementary error profiles.} QuIP\# preserves reasoning and emotional understanding (eq\_bench: +3.61); IQ2\_XXS preserves classification sensitivity (cbd: +11.11). This suggests potential for task-specific method selection.

    \item \textbf{Rotation methods fail at generation.} SpinQuant (B) and ButterflyQuant (C) preserve MC performance but produce incoherent autoregressive text due to missing runtime transforms---an underreported deployment barrier.

    \item \textbf{Cloud-based Polish calibration is practical.} 512-sample Hessians from CulturaX-PL in $\sim$40 minutes on H200 ($\sim$\$2).

    \item \textbf{6.7$\times$ compression enables consumer deployment.} Bielik-11B-v2.3-Instruct from 22\,GB to 3.26\,GB, within reach of 4\,GB consumer GPUs---relevant for edge AI applications.

    \item \textbf{Method-independent quality ceiling.} Four unrelated paradigms (QuIP\#, QTIP, VPTQ, AQLM) converge to 78.1--79.4\% MC accuracy despite different bitrates (2.4--3.58 bpw), suggesting an information-theoretic bound on quality preservation at extreme compression for this model.
\end{enumerate}

\subsection{Future Work}

(1) Complete generative evaluation for Variant F; 0-shot evaluation for all variants; (2) formal Polish vs.\ English calibration ablation; (3) runtime inverse transform implementation; (4) inference speed benchmarking; (5) application to larger Polish models; (6) investigate whether the $\sim$79\% MC ceiling holds for other Slavic language models.

\section*{Acknowledgments}

We thank Krzysztof Wr\'{o}bel and the SpeakLeash project for eq\_bench evaluation, access to the Open Polish LLM Leaderboard harness, and feedback on evaluation methodology. Bielik was developed by SpeakLeash with ACK Cyfronet AGH (grant PLG/2024/016951). We also thank Marcin D\k{a}browski and Bartosz Ba\'{n}kowski for valuable consultations. Cloud infrastructure provided by vast.ai. Total compute budget: $\sim$\$285 from personal funds.

\bibliographystyle{unsrt}

\end{document}